\documentclass[letterpaper, 10 pt, conference]{ieeeconf}  

\usepackage{amsmath}
\makeatletter
\def\maketag@@@#1{\hbox{\m@th\normalfont\normalsize#1}}
\makeatother
\usepackage{amsfonts}

\usepackage{amsthm}
\usepackage[ruled,vlined]{algorithm2e}
\usepackage{mathtools}
\usepackage{tabularx}
\usepackage{graphicx}
\usepackage{subfigure}
\usepackage{enumerate}
\usepackage{stfloats}
\usepackage{float}
\usepackage{url}
\usepackage{verbatim}
\usepackage[linkcolor=black,citecolor=black,urlcolor=black,colorlinks=true]{hyperref}
\usepackage{cite}
\usepackage{hyperref}
\usepackage{algorithmicx}
\usepackage{array}
\usepackage{multirow}
\usepackage{longtable}
\usepackage{rotating}
\usepackage{caption}
\usepackage{graphicx}
\usepackage{booktabs}
\usepackage{array}
\usepackage{ragged2e}
\usepackage{bm}
\usepackage{lipsum,mwe,cuted}
\usepackage{afterpage}

\hypersetup{
	colorlinks=true,
	linkcolor=black,
	urlcolor=blue,
	citecolor=green
}

\DeclareCaptionLabelFormat{fig}{Fig. #2}
\graphicspath{{}}
\IEEEoverridecommandlockouts
\begin{document}
	
	\title{\LARGE \bf
		FLOAT Drone: A Fully-actuated Coaxial Aerial Robot for Close-Proximity Operations
	}
	
	\author{Junxiao Lin, Shuhang Ji, Yuze Wu, Tianyue Wu, Zhichao Han and Fei Gao
	\thanks{All authors are with the State Key Laboratory of Industrial Control Technology, Institute of Cyber-Systems and Control, Zhejiang University, Hangzhou, 310027, China (\textit{Corresponding Author: Fei Gao}).}
	\thanks{E-mail:{\tt\small \{jxlin, fgaoaa\}@zju.edu.cn}}
	}

	\makeatletter
	\let\@oldmaketitle\@maketitle
	\renewcommand{\@maketitle}{\@oldmaketitle
		\includegraphics[width=1.0\linewidth]
		{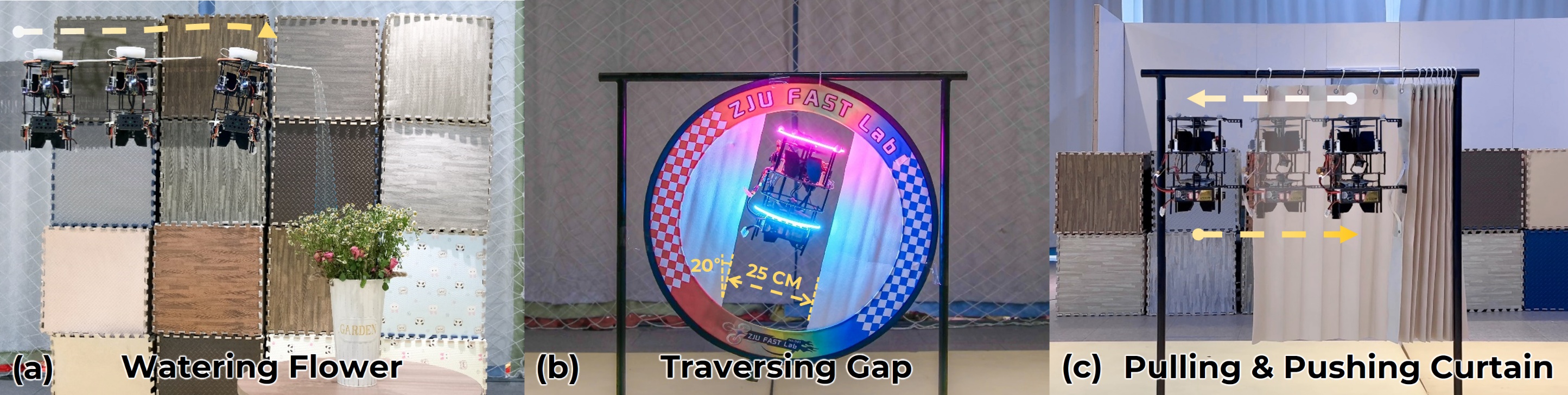}
		\centering
		\captionof{figure}{ \label{fig:top_figure}
		Our proposed FLOAT Drone demonstrates diverse close-proximity operation tasks. (a) Carrying a watering can and watering flowers via a tilted hovering posture. (b) Traversing a narrow gap inclined at 20 degrees, with a width of 25 cm. (c) Pulling and pushing a deformable curtain within a minimum workspace of only 30 cm. Videos on the \href{https://junxiaolin.github.io/float-drone.github.io/}{project website}.
		}
		\vspace{0.0cm} 
	}
	\makeatother
	\maketitle
	\setcounter{figure}{1}
	\thispagestyle{empty}
	\pagestyle{empty}

	\begin{abstract}
	How to endow aerial robots with the ability to operate in close proximity remains an open problem. The core challenges lie in the propulsion system's dual-task requirement: generating manipulation forces while simultaneously counteracting gravity. These competing demands create dynamic coupling effects during physical interactions. Furthermore, rotor-induced airflow disturbances critically undermine operational reliability. Although fully-actuated unmanned aerial vehicles (UAVs) alleviate dynamic coupling effects via six-degree-of-freedom (6-DoF) force-torque decoupling, existing implementations fail to address the aerodynamic interference between drones and environments. They also suffer from oversized designs, which compromise maneuverability and limit their applications in various operational scenarios. To address these limitations, we present FLOAT Drone (FuLly-actuated cOaxial Aerial roboT), a novel fully-actuated UAV featuring two key structural innovations. By integrating control surfaces into fully-actuated systems for the first time, we significantly suppress lateral airflow disturbances during operations. Furthermore, a coaxial dual-rotor configuration enables a compact size while maintaining high hovering efficiency. Through dynamic modeling, we have developed hierarchical position and attitude controllers that support both fully-actuated and underactuated modes. Experimental validation through comprehensive real-world experiments confirms the system's functional capabilities in close-proximity operations.
	\end{abstract}
	
\section{Introduction}
\label{sec:intro}
With advances in system integration and intelligence levels of UAVs, their applications have expanded from environmental observation \cite{zhang2023autofilmer,zhou2023racer,aucone2023dna}, e.g., aerial photography, search and rescue, and environmental monitoring, to missions demanding physical interaction \cite{liu2024compact,wu2023ringrotor,guo2023tactile}, such as object manipulation, transportation, and contact-based infrastructure inspection. 

This shift toward physical interaction necessitates close-proximity operations—defined as precise physical interactions near target objects. Unlike ground robots that inherently exploit ground contact for stabilization \cite{sleiman2023legsr,xiong2024openworld,minniti2019wbmpc}, aerial robots face unique challenges. Drones must simultaneously apply manipulation forces and counteract gravity via their propulsion units. This dynamic coupling effect fundamentally compromises operational stability during physical interactions. Furthermore, the airflow generated by rotors interacts with surrounding environments or manipulated objects, creating complex aerodynamic disturbances that significantly degrade operational reliability during close-proximity tasks. Consequently, to achieve robust close-proximity operations, aerial platforms require solutions that both minimize the dynamic coupling and mitigate airflow interference.

Conventional underactuated UAVs are fundamentally limited in close-proximity operations as they can only generate thrust perpendicular to the airframe plane, inherently coupling translational and rotational motion \cite{underactuated}. This limitation forces continuous attitude adjustments to redirect total thrust for gravity compensation and operational force generation \cite{byun2023icra,lee2021pushdoor,wu2024wholebodycontrolnarrowgaps}. In contrast, fully-actuated platforms achieve 6-DoF force-torque decoupling via specialized actuation systems \cite{bodie2021fulltro,guo2024calligrapher,su2023sequent}. Such capability enables direct generation of multidimensional wrenches without body reorientation, theoretically providing ideal conditions for proximal operations. Nevertheless, existing fully-actuated designs exhibit critical limitations. When applying horizontal forces, the resulting sideways airflow may disturb the target object, thereby impairing operational stability. Moreover, oversized configurations in existing platforms not only compromise maneuverability, but also restrict their applicability in diverse scenarios. These challenges urgently demand novel fully-actuated UAV designs that integrate operational stability, minimized aerodynamic interference, and scenario adaptability for reliable close-proximity operations.

In this paper, we propose the FLOAT Drone, a novel fully-actuated coaxial UAV designed for reliable close-proximity operations. Our core innovation lies in the use of control surfaces that exploit the Bernoulli principle to generate horizontal lift. This design not only mitigates lateral airflow disturbances, but also generates substantial horizontal lift. To the best of our knowledge, this is the first time that control surfaces have been successfully integrated into a fully-actuated UAV system. Additionally,the FLOAT Drone employs a coaxial dual-rotor design, achieving a compact size while maintaining high hovering efficiency. Through dynamic modeling, we develop hierarchical position and attitude controllers enabling seamless transitions between fully-actuated and underactuated modes.

To validate the functional capabilities of FLOAT Drone, we conducted comprehensive real-world experiments. In the experiment involving pulling and pushing a deformable curtain, the drone demonstrates significant suppression of lateral airflow interference compared to conventional UAVs. By traversing an inclined narrow gap and executing flower watering tasks with tilted hovering, the drone verifies its compact size and full actuation capabilities. Bi-modal flight test and continuous hovering attitude tracking experiment highlight the precise control performance of the drone.  

The main contributions of this paper are as follows:
\begin{itemize}
	\item We propose a novel fully-actuated drone design suitable for close-proximity operations. Through innovative control surface and coaxial dual-rotor design, we achieve a low-airflow-interference and compact flying platform;
	\item By thoroughly analyzing the dynamics of the proposed drone, we develop the hierarchical controller, enabling flexible switching between two modes;
	\item Through extensive real-world experiments, we validate the functionality of FLOAT Drone in close-proximity operations.
\end{itemize}

\section{Related Work}
Fully-actuated UAVs can be broadly classified into two categories: fixed-tilt configurations and variable-tilt configurations \cite{rashad2020review}.

\subsection{Fixed-Tilt Fully Actuated UAV}
\label{sec:2A}
Fixed-tilt configurations achieve full actuation by presetting specific tilt angles for each rotor, enabling multirotors to generate thrust in multiple directions. This approach contrasts with underactuated UAVs, where rotors are aligned parallel and can only produce unidirectional thrust. For instance, \cite{rajappa2015hexa} builds upon the traditional hexacopter framework, where each rotor is tilted at a fixed angle, allowing the UAV to generate forces in any direction. \cite{tognon2018seven} uses seven rotors, each producing positive thrust, with varying spatial orientations, all arranged within a single horizontal plane to achieve full actuation. \cite{brescianini2016omni} features the first fully-actuated octocopter capable of omnidirectional flight, with eight rotors fixed at the vertices of a cube. \cite{park2018odar} introduces a beam-shaped omnidirectional octocopter, specifically designed for aerial interaction tasks. 

The primary advantage of fixed-tilt configurations lies in their ability to achieve full actuation while maintaining a relatively simple mechanical structure and control system, similar to underactuated UAVs. However, due to the non-vertical alignment of the rotors, additional thrust must be used to counteract internal forces during hover, resulting in lower energy efficiency when compared to conventional parallel rotor configurations.

\subsection{Variable-Tilt Fully Actuated UAV}
Variable-tilt configurations achieve full actuation by dynamically adjusting the orientation of rotors, thereby changing the direction of the thrust vector. For example,\cite{markus2015quad} modifies the traditional quadcopter by equipping each rotor with a tilting servo, allowing it to rotate around an axis collinear with the arm, with a maximum tilt angle of 25°. \cite{kamel2018voliro} incorporates six independent tilting servos, one per rotor, based on a hexacopter framework. This design is recognized as the first variable-tilt configuration capable of omnidirectional flight. \cite{zheng2020tiltdrone} introduces a novel gimbal mechanism into a quadcopter design, wherein two servos are used to synchronize the tilting of four rotors within the roll and pitch planes, achieving a maximum tilt angle of 30°. \cite{paulos2018emul} employs swashplateless rotors to generate multidirectional thrust through periodic modulation of motor torque. This configuration achieves full actuation with only two rotors, although the maximum tilt angle is 8° and the high-frequency torque modulation can cause motor overheating. 

The advantage of variable-tilt configurations is their ability to switch between fully actuated and underactuated modes based on mission requirements, while also maintaining high energy efficiency during hover. However, these designs tend to introduce increased complexity in terms of both structural components and control systems, along with additional size and weight due to the tilting mechanisms.

\section{Design}
\label{sec:3}
\subsection{Compactness Analysis of Rotor Configurations}

\begin{figure}[t]   
	\centering
	{\includegraphics[width=1.0\columnwidth]{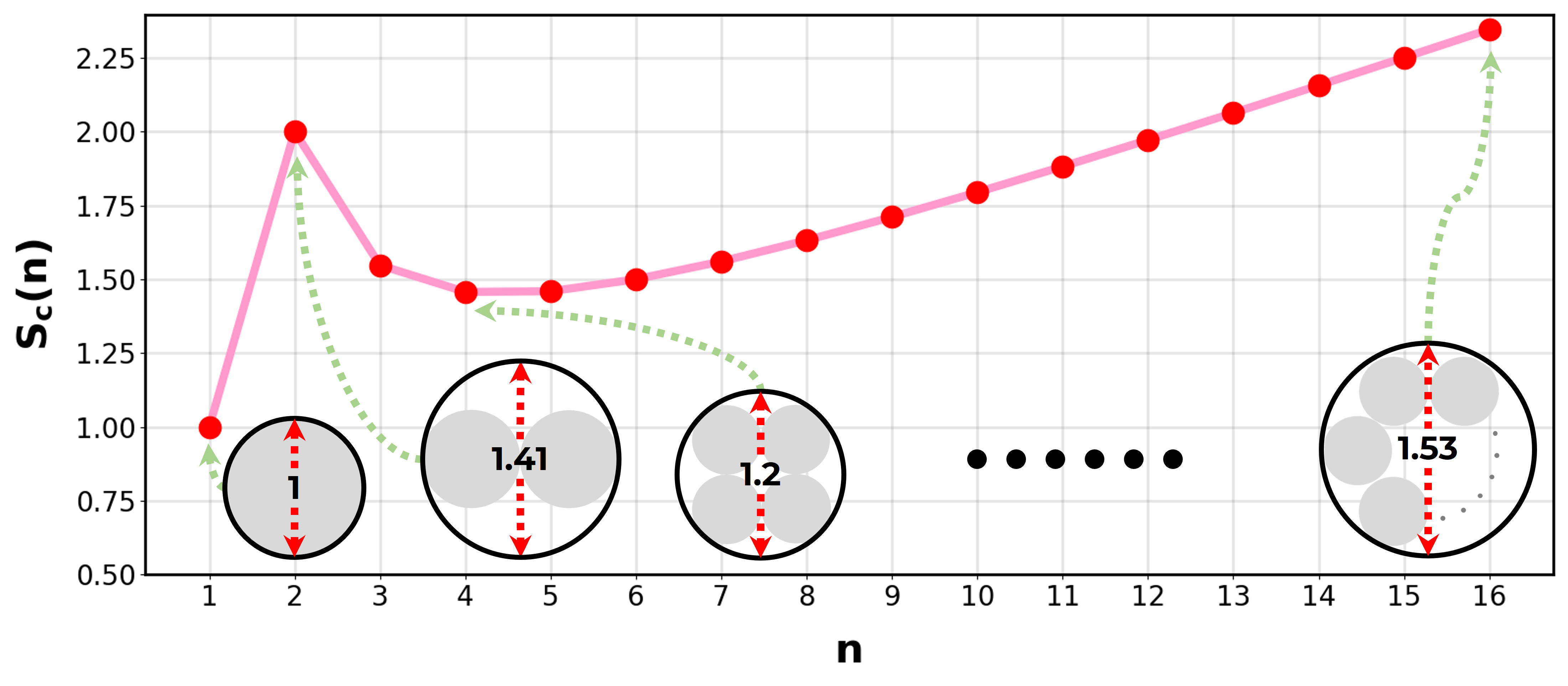}}
	\captionsetup{font={small}}
	\caption{ \label{fig:S(n)}Visualization of $S_{c}(n)$. Y-axis represents the area of the circumscribed circle for different numbers of rotors, with the single rotor as the baseline. The values within circumscribed circles represent the diameter ratio of the circumscribed circle.}
	\vspace{-0.6cm} 
\end{figure}

\begin{figure*}[t]
	\centering
	{\includegraphics[width=0.9\textwidth]{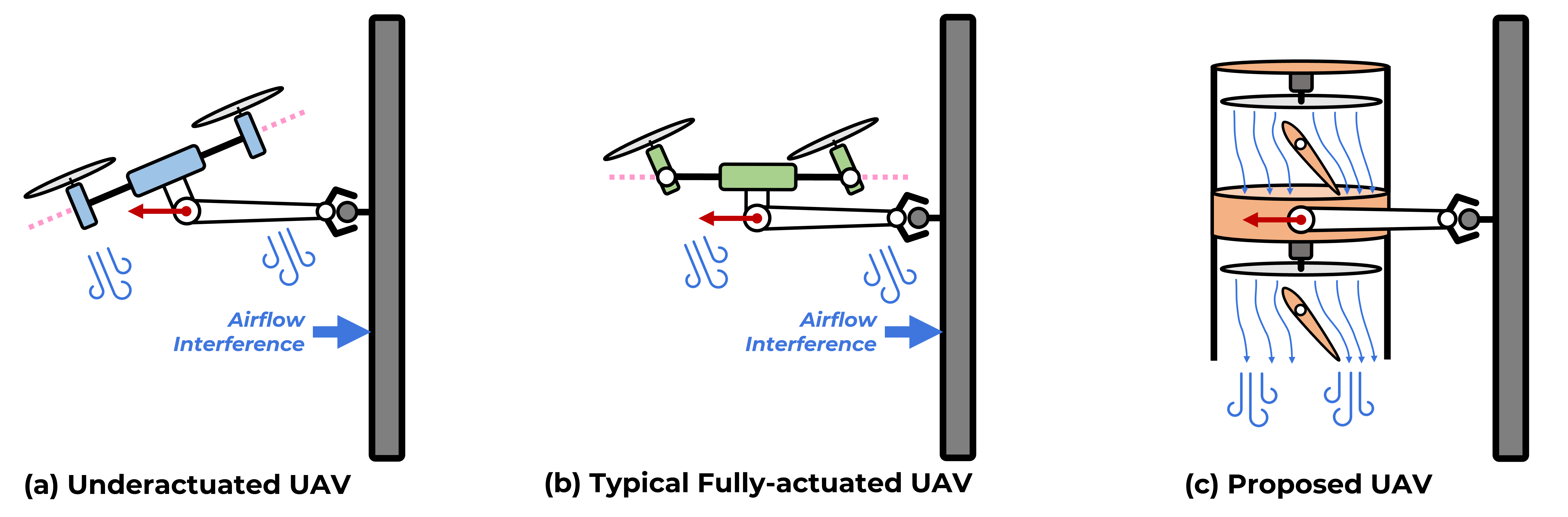}}
	\captionsetup{font={small}}
	\caption{ \label{fig:horizontal_force_large} Comparison of horizontal force generation among different UAV configurations. The red arrows denote the direction of the horizontal thrust generated by the drone, while the blue arrows indicate the direction of the forces exerted on the object by the downwash airflow.}
	\vspace{-0.5cm}
\end{figure*}

\begin{figure}[t]
	\centering
	{\includegraphics[width=1.0\columnwidth]{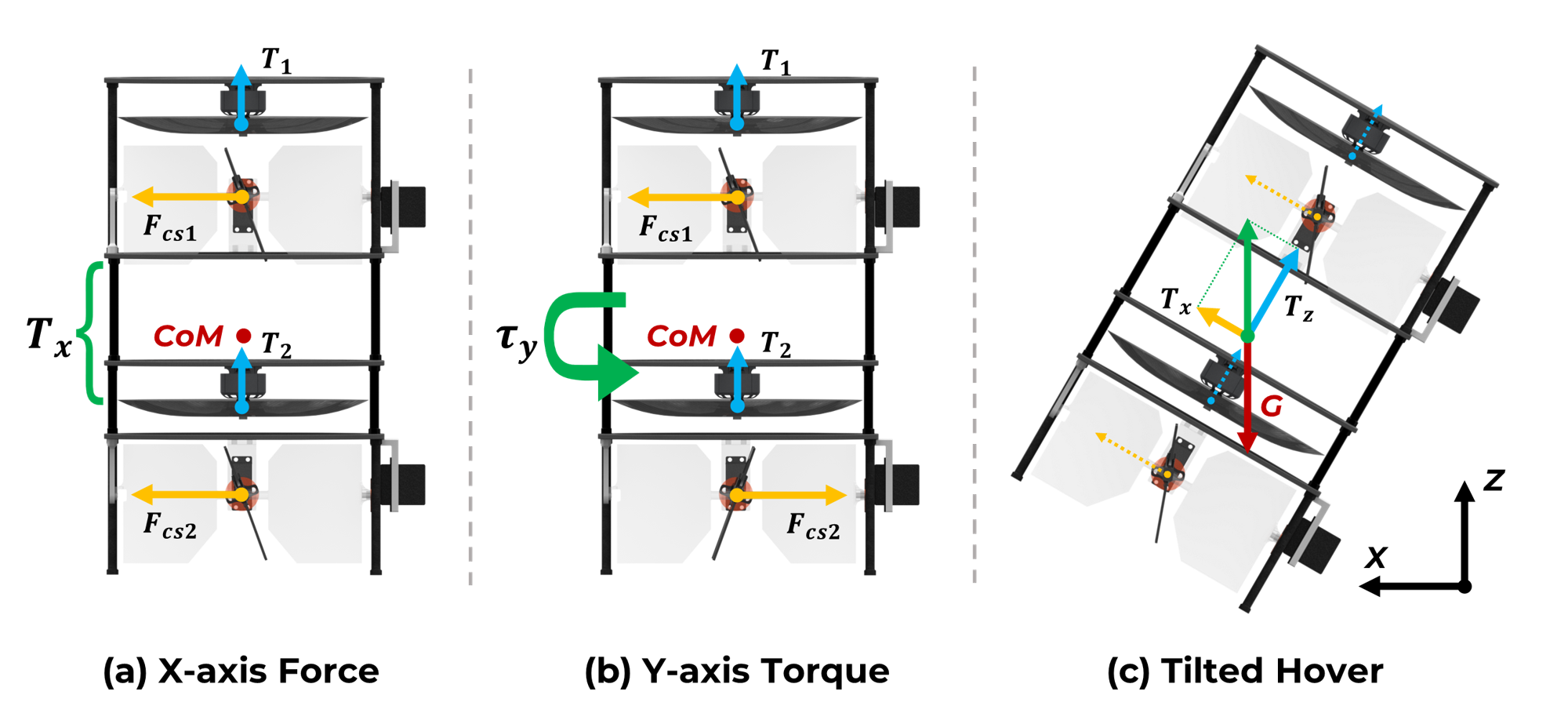}}
	\captionsetup{font={small}}
	\caption{ \label{fig:fully_actuation_implementation}Schematic diagram for the implementation of full actuation. $T_1$ and $T_2$ represent the thrusts generated by the upper and lower rotors, respectively. $T_z$ is the total thrust in the vertical direction of the drone. $F_{cs1}$ and $F_{cs2}$ denote the lift forces produced by the upper and lower control surfaces, respectively. $T_x$ and $\tau_y$ are the total force and torque in the horizontal direction of the UAV. $G$ represents the gravity of the drone.}
	\vspace{-0.5cm} 
\end{figure}

As previously mentioned, to enhance the maneuverability of UAVs and expand their applications, minimizing the size of drones while maintaining flight efficiency is critically important. In this section, we analyze the compactness of different rotor configurations under the same hovering efficiency.

In practical applications, the size of a drone is typically characterized by the area of its minimum circumscribed circle or sphere, which also serves as the basis of most obstacle avoidance algorithms \cite{zhou2021egoplanner}. Therefore, we select the area of the minimum circumscribed circle of the UAV’s horizontal projection as its valid size.

Based on the analysis from previous work \cite{qin2020gemini} \cite{lin2024skater}, we know that the ideal power $P$ of a rotor to generate lift $F$ is
\begin{equation}
	\label{eqn_P}
	P(F) = \sqrt{\frac{F^3}{2\pi r^2\rho } } ,
\end{equation}
where $r$ is the radius of the rotor, and $\rho $ is the air density.

Then, we define the hovering efficiency $\eta _{h} $ of a UAV as
\begin{equation}
	\label{eqn_eta}
	\eta _{h}(n) =\frac{m}{n\times P(mg/n)} = \frac{r\sqrt{2\pi n \rho } }{g\sqrt{mg} } ,
\end{equation}
where $m$ is the mass of the drone, $n$ is the number of rotors, and $g$ is the gravitational acceleration.

Through straightforward analysis and calculation\cite{pan2023canfly}, we can derive the radius of the circumscribed circle $R$ for the rotorcraft UAV as
\begin{equation}
	\label{eqn_R}
	R = \left\{\begin{matrix}
		r,  & n = 1,\\
		\frac{(1+\sin (\pi / n))}{\sin (\pi / n)}r,   & n\ge 2.
	\end{matrix}\right.
\end{equation}
Then, the area of the minimum circumscribed circle $S_c$ is
\begin{equation}
	\label{eqn_S}
	S_{c}(n) = \left\{\begin{matrix}
		\frac{\eta _{h}^2mg^3}{2\rho}, & n = 1,\\
		\frac{\eta _{h}^2mg^3}{2\rho}\frac{(1+\sin (\pi / n))^2}{n \sin (\pi / n)^2},   & n \ge 2.
	\end{matrix}\right.
\end{equation}

Through the visualization of $S_{c}(n)$ in Fig. \ref{fig:S(n)}, it is evident that under equivalent hovering efficiency and weight conditions, the single-rotor configuration exhibits the minimal size among rotor configurations. Furthermore, experimental evidence from \cite{pan2023canfly} demonstrates that a coaxial dual-rotor configuration can enhance the hovering efficiency without increasing the valid size of the UAV. Consequently, the proposed aerial robot adopts a coaxial dual-rotor architecture.

\subsection{Horizontal Force Generation Strategy}
For close-proximity operations, minimized aerodynamic interference is crucial, which presents challenges in design configurations. Fig. \ref{fig:horizontal_force_large} illustrates a typical close-range task: pulling an object while approaching the target, where the drone needs to exert a force away from the target.

For underactuated multirotors, tilting the attitude can generate pulling force. However, this results in the downwash airflow from the rotors acting on the object, producing a counteracting force that hinders task completion, as shown in Fig. \ref{fig:horizontal_force_large}(a). For conventional fully-actuated UAVs, although tilting the attitude is unnecessary, the rotor airflow inevitably blows towards the object, similarly interfering with task execution, as shown in Fig. \ref{fig:horizontal_force_large}(b).

To mitigate this issue, we propose the use of control surfaces to provide horizontal force, as depicted in Fig. \ref{fig:horizontal_force_large}(c). When the airflow beneath the rotors passes through control surfaces with a certain attack angle, a velocity difference is created between the two sides of the control surfaces. According to Bernoulli's principle, this velocity difference generates a pressure differential, thereby exerting a horizontal force on the drone. This force, derived from the pressure differential, does not produce significant airflow in the horizontal direction, effectively alleviating the airflow interference between the UAV and the object during proximal operations. Subsequent experiments have validated the efficacy of this approach.
\begin{figure}[t]   
	\centering
	{\includegraphics[width=1.0\columnwidth]{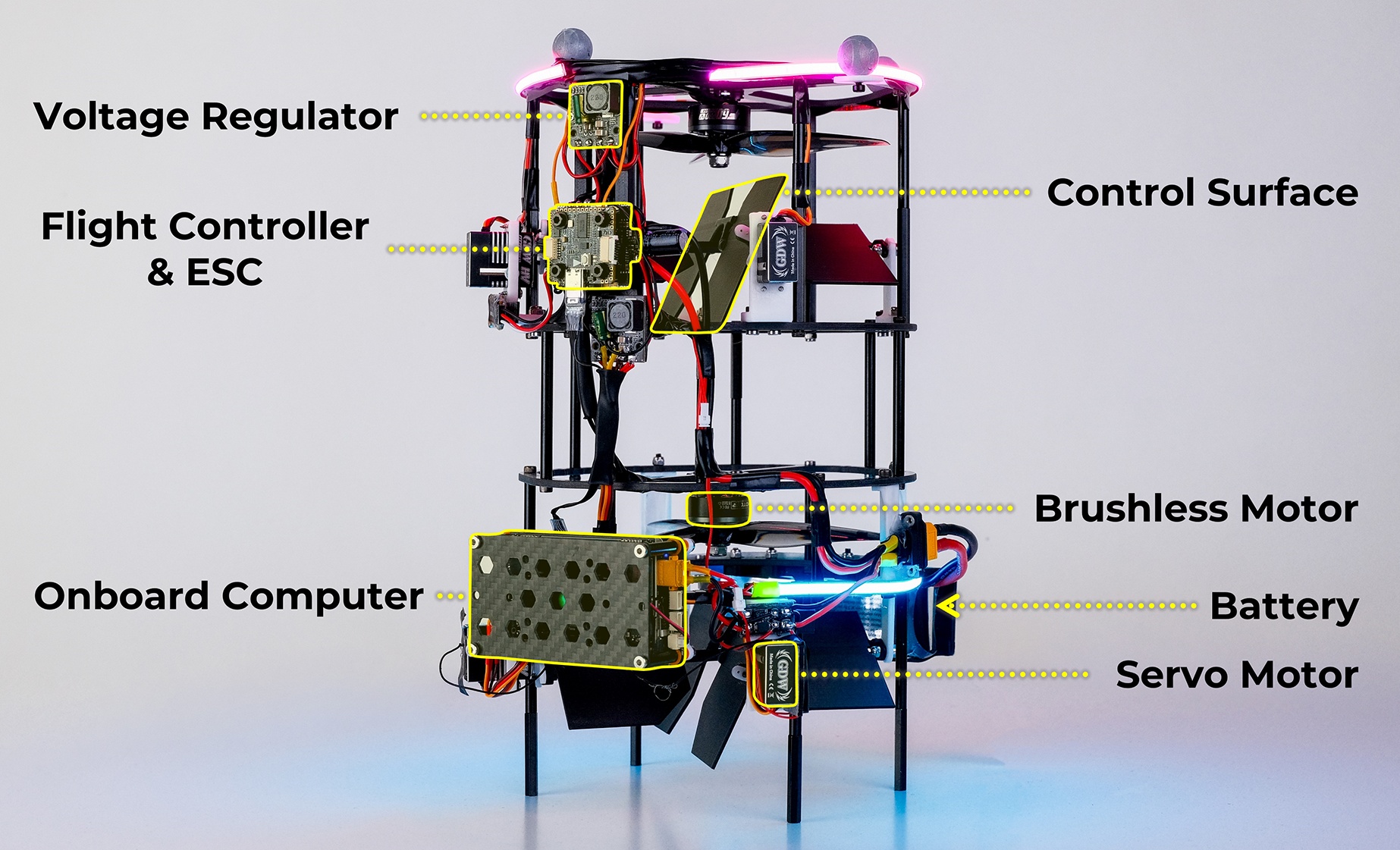}}
	\captionsetup{font={small}}
	\caption{ \label{fig:hardware}Prototype and hardware configuration of FLOAT Drone.}
	\vspace{-0.2cm} 
\end{figure}

\begin{table}[t]
	\footnotesize
	\renewcommand{\arraystretch}{1.1}
	\setlength{\tabcolsep}{3pt}
	\centering
	\caption{Component Configuration and Weight Distribution}
	\label{tab:component_weights}
	\noindent
	\resizebox{\linewidth}{!}{
		\begin{tabular}{@{} 
				>{\raggedright\arraybackslash}p{2.3cm} |
				>{\raggedright\arraybackslash}p{3.2cm} | 
				c  |
				r  |
				r 
				@{}}
			\toprule[2pt]
			\textbf{Component} & 
			\textbf{Model} & 
			\textbf{Units} & 
			\textbf{Weight (g)} & 
			\textbf{Weight\%} \\
			\midrule
			Battery & TATTU FPV 2300mAh 4S & 1 & 236 & 24.8 \\
			OnBoard Computer & NVIDIA Jetson Orin NX & 1 & 81 & 8.5 \\
			Servo Motor & GDW DS290MG & 4 & 80 & 8.4 \\
			Brushless Motor & T-Motor F60Pro & 2 & 66 & 6.9 \\
			Control Surfaces & 3D Printed Parts & 4 & 66 & 6.9 \\
			Propellers & Gemfan 6045-3 6-inch & 2 & 14 & 1.5 \\
			ESC & HAKRC 65A & 1 & 11 & 1.2 \\
			Flight Controller & Holybro Kakute H7 Mini & 1 & 8 & 0.8 \\
			\bottomrule[2pt]
	\end{tabular}}
	\vspace{-0.5cm} 
\end{table}

\subsection{Full Actuation Implementation}
To achieve the low-disturbance proximal operation described earlier, the UAV needs to generate horizontal forces without body reorientation. This requires the UAV to independently control forces and torques along three axes. As shown in Fig. \ref{fig:fully_actuation_implementation}, this design uses an innovative actuator setup to achieve full actuation: the robot is equipped with six actuators, including two symmetrically placed rotor-control surface units (each with one rotor and two servo-controlled surfaces), located above and below the center of mass (CoM).

For vertical (Z-axis) control, the two rotors produce thrust and torque by rotating at different speeds. When they rotate at the same speed, they generate pure vertical thrust $ T_z = T_1 + T_2$. When they rotate at different speeds, they create yaw torque. For horizontal (X/Y-axis) control, the control surfaces work together—when the upper and lower surfaces move in the same direction, they produce combined aerodynamic forces for translational thrust $T_x = F_{cs1} + F_{cs2}$. When they move in opposite directions, they generate pure roll or pitch torques $\tau_y$ through the couple effect, as shown in Fig. \ref{fig:fully_actuation_implementation}(a) and Fig. \ref{fig:fully_actuation_implementation}(b). This ability to independently control forces and moments along three axes allows the UAV to adjust its hovering attitude, as illustrated in Fig. \ref{fig:fully_actuation_implementation}(c). By adjusting the combination of vertical thrust $T_z$ and horizontal lift $T_x$, the UAV can balance the total force with gravity $G$ even when the body is tilted.

This analysis explains how the fully-actuated mechanism works. In the subsequent sections, we will develop the UAV’s dynamic model and design the controller.

\begin{figure}[t]   
	\centering
	{\includegraphics[width=0.7\columnwidth]{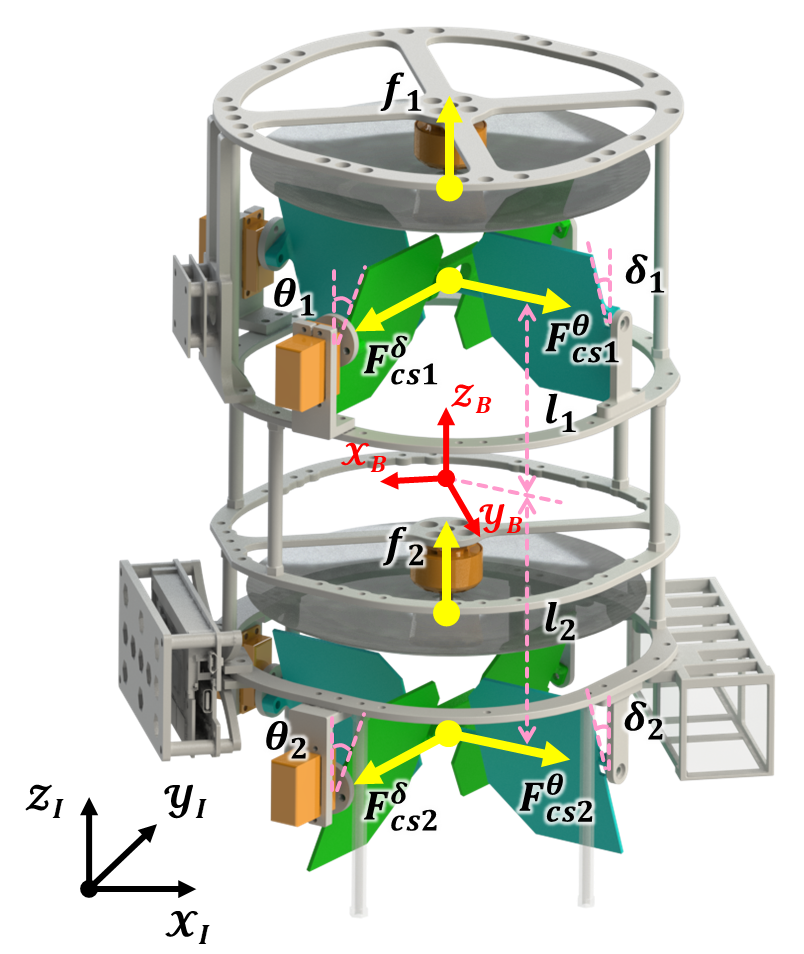}}
	\captionsetup{font={small}}
	\caption{ \label{fig:dynamics} Coordinate definitions and force analysis.}
	\vspace{-0.5cm} 
\end{figure}
\subsection{Hardware Implementation}
The prototype and hardware configuration of FLOAT Drone are shown in Fig. \ref{fig:hardware}. With a compact design measuring 180×230×290 mm, it achieves a minimum circumscribed circle diameter of 250 mm in horizontal projection while maintaining a total weight of 952 g. Table 1 lists the key components with their model specifications and weight distributions.

To optimize airflow through the central channel, all non-propulsion components are arranged in the peripheral ring area, preserving an unobstructed axial flow path. In addition, by optimizing the component layout, we align the drone’s CoM with the rotational axes of the upper and lower motors. This design eliminates pitch and roll torques induced by CoM deviation, significantly enhancing flight stability and simplifying control algorithms.
\section{Dynamics and Control}
\subsection{Dynamics}
First, we define two reference coordinate systems: the inertial reference frame $\bm{\mathcal{F}}_{\mathcal{I} }({\bm{x}_{\mathcal{I} }},\bm{y}_{\mathcal{I} },\bm{z}_{\mathcal{I} })$ and the body-fixed reference frame $\bm{\mathcal{F}}_{\mathcal{B} }({\bm{x}_{\mathcal{B} }},\bm{y}_{\mathcal{B} },\bm{z}_{\mathcal{B} })$. The actuator outputs of the FLOAT Drone are $ \bm{u} = \{f_1, f_2, \theta_1, \theta_2, \delta_1, \delta_2\}$, which represent the thrust of the two rotors and the angles of the four servo motors, respectively. The lift forces generated by the four control surfaces are denoted as $\bm{F_{cs}} = \{F_{cs1}^{\theta}, F_{cs2}^{\theta}, F_{cs1}^{\delta}, F_{cs2}^{\delta}\}$. The distances from the upper and lower control surfaces to the CoM are $l_{1}$ and $l_{2}$, respectively, as shown in Fig. \ref{fig:dynamics}.

From \cite{beard2012small}, the lift equation for the control surface $F_{CS}$ and the thrust generated by the rotor $f$ are as follows:
\begin{equation}
	\label{eqn:F_CS}
	F_{CS} = \frac{1}{2}\rho S C_{L} V_{a}^{2},
\end{equation}
\begin{equation}
	\label{eqn:f}
	f = \frac{1}{2} \rho A (V_{\text{out}}^{2} - V_{\text{in}}^{2}),
\end{equation}
where $\rho$ is the air density, $S$ is the valid area of the control surface, $C_L$ is the lift coefficient, and $V_a$ is the airflow velocity over the control surface; $A$ is the swept area of the propeller, $V_{\text{out}}$ is the airflow velocity exiting the propeller, and $V_{\text{in}}$ is the airflow velocity entering the propeller.

Assuming the drone is in low-speed motion with zero incoming flow velocity at the upper rotor, the thrusts of upper and lower motors can be respectively expressed as:
\begin{equation}
	\label{eqn:f1}
	f_{1} = \frac{1}{2} \rho A V_{1}^{2},
\end{equation}

\begin{equation}
	\label{eqn:f2}
	f_{2} = \frac{1}{2} \rho A (V_{2}^{2} - V_{1}^{2}),
\end{equation}
where $V_1$ and $V_2$ are the airflow velocities over the upper and lower control surface.

Considering the angle of the control surface is relatively small, $C_{L}$ can be approximated as linearly proportional to the angle \cite{beard2012small}: 
\begin{equation}
	\label{eqn:C_L}
	C_{L} = C_{l}\theta,
\end{equation}
 where $C_{l}$ is a proportionality constant. Through straightforward calculations, the relationship between control surface lifts $\bm{F_{cs}}$ and $\bm{u}$ can be derived as follows:
\begin{equation}
	\label{eqn:F_cs}
	\bm{F_{cs}} = \begin{bmatrix} F_{cs1}^{\theta}
		\\ F_{cs2}^{\theta}
		\\ F_{cs1}^{\delta}
		\\ F_{cs2}^{\delta}
		
	\end{bmatrix}
	= \begin{bmatrix} K_{cs} f_{1} \theta_1 
		\\ K_{cs} (f_{1} +f_{2}) \theta_2 
		\\ K_{cs} f_{1} \delta_1 
		\\ K_{cs} (f_{1} +f_{2}) \delta_2 
		
	\end{bmatrix}, K_{cs} = \frac{SC_{l}}{A}.
\end{equation}

According to the Newton-Euler equations, we define the dynamics of the system as
\begin{equation}
	\label{eqn:p_ddot}
	m \bm{\ddot{p}} = m \bm{g} + \bm{R}\bm{T_{B}},
\end{equation}
\begin{equation}
	\label{eqn:omega_dot}
	 \bm{J} \bm{\dot{\omega}} = -\bm{\omega}  \times \bm{J} \bm{\omega}  + \bm{\tau _B},
\end{equation}
Where \( m \) represents the mass of the robot, \( \bm{p} \) is the position vector of the robot in $\bm{\mathcal{F}}_{\mathcal{I} }$, $\bm{g} = [0,0,-9.81 \ \mathrm{m/s^2}]^T$ is the gravity vector, \( \bm{R} \) is the attitude rotation matrix  from $\bm{\mathcal{F}}_{\mathcal{B} }$ to $\bm{\mathcal{F}}_{\mathcal{I} }$, and $ \bm{T_B} = [{T}_{x}, {T}_{y}, {T}_{z}]^{T}$ is the thrust generated by the system  in $\bm{\mathcal{F}}_{\mathcal{B} }$:
\begin{equation}
	\label{eqn:T_B}
	\bm{T}_B
	= \begin{bmatrix}
		\frac{\sqrt{2} }{2} (-F_{cs1}^{\theta} - F_{cs2}^{\theta} + F_{cs1}^{\delta} + F_{cs2}^{\delta})
		\\ 
		\frac{\sqrt{2} }{2} (F_{cs1}^{\theta} + F_{cs2}^{\theta} + F_{cs1}^{\delta} + F_{cs2}^{\delta})
		\\ 
		f_1 + f_2
	\end{bmatrix}.
\end{equation}

\( \mathbf{J} \) represents the inertia of the robot, \( \boldsymbol{\omega} \) is the angular velocity in $\bm{\mathcal{F}}_{\mathcal{B} }$, and $ \boldsymbol{\tau_B}  = [{\tau}_{x}, {\tau}_{y}, {\tau}_{z}]^{T} $ is the torque generated by the system in $\bm{\mathcal{F}}_{\mathcal{B} }$:
\begin{equation}
	\label{eqn:tau_B}
	\bm{\tau_B} 
	= \begin{bmatrix}
		\frac{\sqrt{2} }{2} (-(F_{cs1}^{\theta} + F_{cs1}^{\delta})l_1 + (F_{cs2}^{\theta} + F_{cs2}^{\delta})l_2)
		\\ 
		\frac{\sqrt{2} }{2} ((-F_{cs1}^{\theta} + F_{cs1}^{\delta})l_1 - (-F_{cs2}^{\theta} + F_{cs2}^{\delta})l_2)
		\\ 
		\frac{k_M }{k_F}(f_1 - f_2)
	\end{bmatrix},
\end{equation}
where $k_M$ and $k_F$ are the rotor torque and thrust coefficient. Based on (\ref{eqn:F_cs}),(\ref{eqn:T_B}) and (\ref{eqn:tau_B}), the mixing relationship between the actuator output $\bm{u}$ of the FLOAT Drone and the system control inputs $\bm{T_B}$ and $\bm{\tau_B}$ can be expressed as follows:
\begin{equation}
	\label{eqn:mixer}
	\left\{
	\begin{aligned}
		f_1 &= \frac{1}{2}\left(T_z + \frac{k_F}{k_M} \tau_z\right), \\
		f_2 &= \frac{1}{2}\left(T_z - \frac{k_F}{k_M} \tau_z\right), \\
		\theta_1 &= \frac{\sqrt{2}(-T_x l_2 + T_y l_2 - \tau_X - \tau_y)}{K_{cs}(l_1 + l_2)\left(T_z + \frac{k_F}{k_M} \tau_z\right)}, \\
		\theta_2 &= \frac{\sqrt{2}(-T_x l_1 + T_y l_1 + \tau_X + \tau_y)}{2K_{cs}(l_1 + l_2)T_z}, \\
		\delta_1 &= \frac{\sqrt{2}(T_x l_2 + T_y l_2 - \tau_X + \tau_y)}{K_{cs}(l_1 + l_2)\left(T_z + \frac{k_F}{k_M} \tau_z\right)}, \\
		\delta_2 &= \frac{\sqrt{2}(T_x l_1 + T_y l_1 + \tau_X - \tau_y)}{2K_{cs}(l_1 + l_2)T_z}.
	\end{aligned}
	\right.
\end{equation}

\begin{figure}[t]   
	\centering
	{\includegraphics[width=1.0\columnwidth]{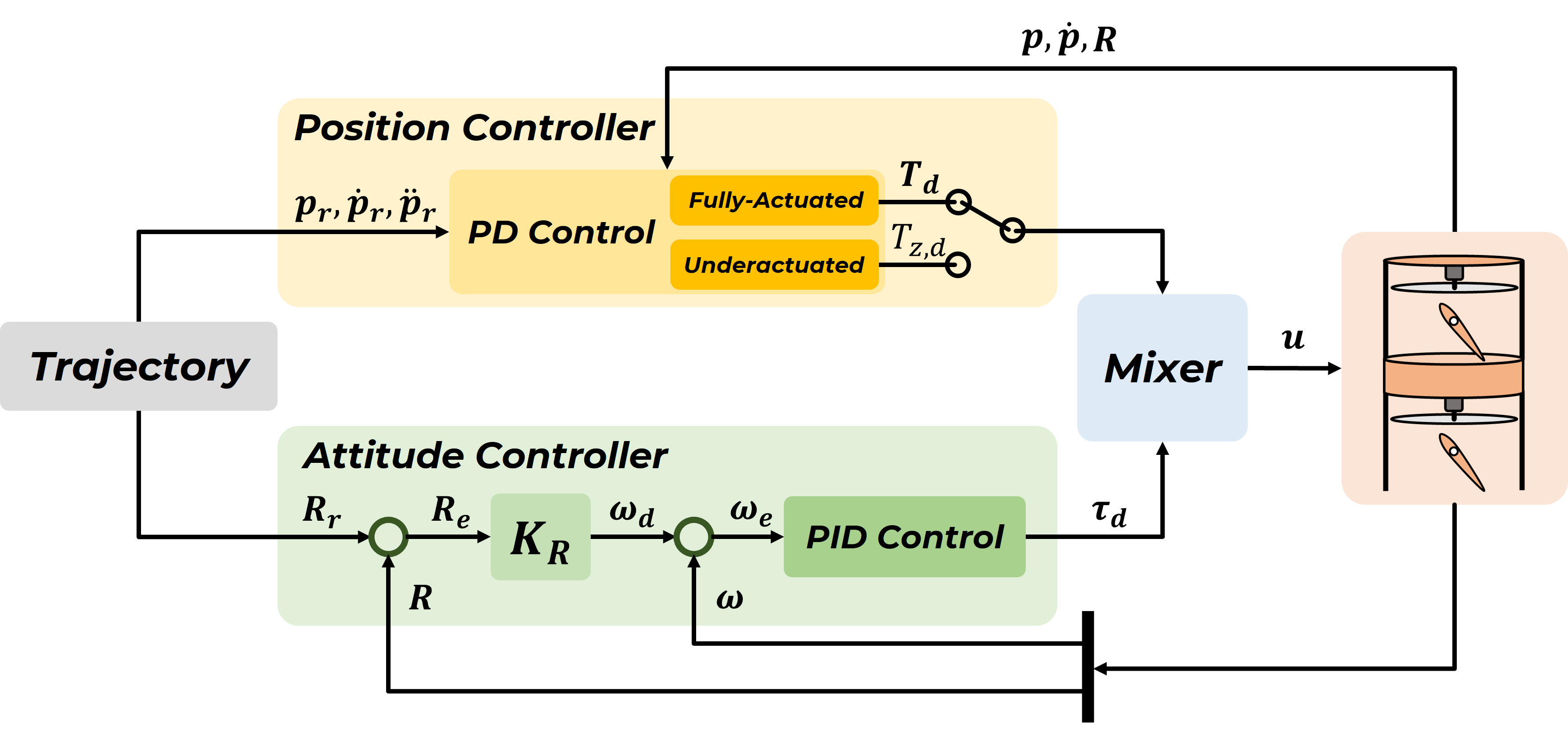}}
	\captionsetup{font={small}}
	\caption{ \label{fig:control_framework} The control framework of the propose system.}
	\vspace{-0.5cm} 
\end{figure}

\begin{figure*}[t]   
	\centering
	{\includegraphics[width=1.95\columnwidth]{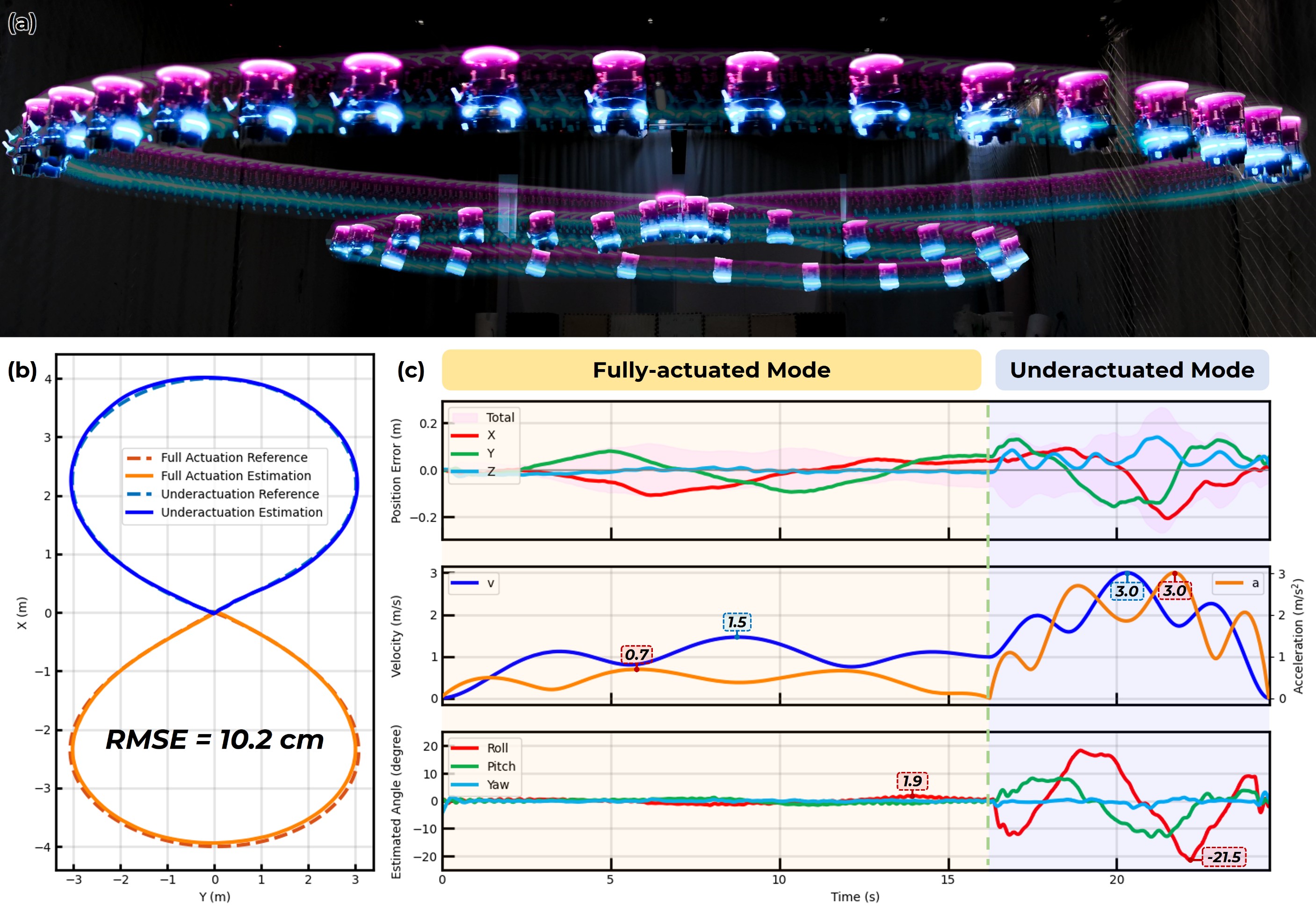}}
	\captionsetup{font={small}}
	\caption{ \label{fig:traj_track}Real-world trajectory tracking experiments and the corresponding experimental data. (a) Snapshot of bi-modal trajectory tracking test. (b) Visualized position tracking curve. (c) State curves of FLOAT Drone during the experiment, including position error, desired velocity and acceleration, and estimated attitude angle.}
	\vspace{-0.2cm} 
\end{figure*}
\subsection{Position Control}
The control framework of the FLOAT Drone is shown in Fig. \ref{fig:control_framework}, which primarily consists of two parts: the position controller and the attitude controller. The position controller is implemented as a cascaded PD controller. First, the desired acceleration $\bm{\ddot{p}_{d}} $ is calculated as follows:
\begin{equation}
	\label{eqn:p_ddot_des}
	\bm{\ddot{p}_{d}} = \bm{\ddot{p}_{r}} + \bm{K_{v}}(\bm{\dot{p}_{r}} + \bm{K_{p}}(\bm{p_{r}} - \bm{p}) - \bm{\dot{p}}),
\end{equation}
where $\bm{\ddot{p}_{r}}$, $\bm{\dot{p}_{r}}$ and $\bm{{p}_{r}}$ denote the reference acceleration, velocity and position obtained from reference trajectory; $\bm{\dot{p}}$ and $\bm{{p}}$ represent the measured velocity and position of the drone; and $\bm{K_{v}}$ and $\bm{K_{p}}$ are positive definite gain matrices.

When the drone operates in fully-actuated mode, its three-dimensional desired thrust $\bm{T_d}$ can be expressed as
\begin{equation}
	\label{eqn:T_des}
	\bm{T_d} = \bm{R}^T (m \bm{\ddot{p}_{d}} - m \bm{g} ).
\end{equation}

However, in underactuated mode, the drone can only generate thrust along $\bm{z}_{\mathcal{B}}$, which means that $T_x$ and $T_y$ are both zero. In this case, the desired collective thrust $T_{z,d}$ can be expressed as
\begin{equation}
	\label{eqn:T_z_des}
	T_{z,d} = (m \bm{\ddot{p}_{d}} - m \bm{g} ) \cdot \bm{z}_{\mathcal{B}} . 
\end{equation}

\subsection{Attitude Control}
The attitude controller of FLOAT Drone is a cascaded PID controller. First, we define the attitude error $\bm{R_e}$ as
\begin{equation}
	\label{eqn:R_e}
	\bm{R_e} = \frac{1}{2}(\bm{R_{r}^{T}} \bm{R} - \bm{R^{T}} \bm{R_r})^{\vee} , 
\end{equation}
where $\bm{R_r}$ and $\bm{R}$ denote the reference and measured attitude rotation matrices, respectively, and \( (\cdot)^\vee \) represents the mapping of elements from the Lie algebra \( \mathfrak{so}(3) \) to vectors in \( \mathbb{R}^3 \). For the fully-actuated mode, $\bm{R_r}$ can be arbitrarily specified by the user, while in the underactuated mode, $\bm{R_r}$ needs to be derived from the reference trajectory \cite{mellinger2011minimumsnap}.

The desired angular velocity $\bm{\omega_d}$ and angular velocity error $\bm{\omega_e}$ are defined as follows:
\begin{equation}
	\label{eqn:omega_d}
	\bm{\omega_d} =  \bm{K_{R}} \cdot \bm{R_e}, 
\end{equation}
\begin{equation}
	\label{eqn:omega_e}
	\bm{\omega_e} = \bm{\omega_d} - \bm{\omega}, 
\end{equation}
where $\bm{K_{R}}$ is a positive definite gain matrix, and $\bm{\omega}$ represents the measured angular velocity.

Then, the desired torque $\bm{\tau_d}$ is calculated using positive definite PID gain matrices $\{ \bm{K_{P,\omega}}, \bm{K_{I,\omega}}, \bm{K_{D,\omega}}\}$:
\begin{equation}
	\label{eqn:tau_d}
	\bm{\tau_d} = \bm{K_{P,\omega}} \bm{\omega_e} + \bm{K_{I,\omega}} \int \bm{\omega_e} + \bm{K_{D,\omega}} {\bm{\dot{\omega}_e}}. 
\end{equation}

\section{Experiments}
\subsection{Seamless Mode Transition in Trajectory Tracking}
In all subsequent experiments, the robot's position and pose estimation are obtained from a NOKOV motion capture system\footnote{https://www.nokov.com}. The FLOAT Drone is capable of operating in both fully-actuated and underactuated flight modes, achieving a balance between flying performance and energy efficiency. The underactuated mode can only activate four actuators, resulting in energy savings compared to full actuation mode. Additionally, the simplified control framework enables easier implementation of high-speed flight. On the other hand, while the fully-actuated mode requires more complex planning and control algorithms, its decoupled thrust vector characteristics provide higher control precision for close-range operational tasks. 

In this experiment, the drone achieves seamless dual-mode switching control during trajectory tracking, as shown in Fig.\ref{fig:traj_track}(a). In the experimental setup, the FLOAT Drone is required to track a two-dimensional 8-shaped trajectory, and different motion metrics are adopted for the two operational modes, as shown in Fig.\ref{fig:traj_track}(b). In the fully-actuated mode, the system sets the trajectory with a maximum velocity of 1.5 m/s and a maximum acceleration of 0.7 $\mathrm{m/s^2}$, while the three-axis attitude angles were all set to zero. When switching to the underactuated mode, the trajectory’s maximum velocity was extended to 3.0 m/s, and the acceleration limit was increased to 3.0 $\mathrm{m/s^2}$, with the desired attitude calculated from the trajectory using differential flatness. The root-mean-square error (RMSE) is adopted as the evaluation metric for trajectory tracking performance, defined mathematically as:
\begin{equation}
	\label{eqn:rmse}
	RMSE = \sqrt{\frac{\sum_{i=0}^{N} \left \| \bm{p}_r(i)-\bm{p}(i) \right \|^2 }{N} },
\end{equation}
where $\bm{p}_r(i)$ and $\bm{p}(i)$ are the $i^{th}$ sampled reference and estimated positions of the vehicle, respectively.

Experimental data are shown in Fig.\ref{fig:traj_track}(c). By analyzing the data, the drone achieves a RMSE of 10.2 cm in position tracking, particularly in the fully-actuated mode, where the maximum attitude tilt angle was strictly controlled within 2 degrees. This achievement not only validated the precise trajectory tracking capability of the designed control system but also successfully demonstrated the drone’s potential to dynamically adjust its mode according to task requirements.

\subsection{Continuous Hovering Attitude Tracking Control}
\begin{figure}[t]   
	\centering
	{\includegraphics[width=1.0\columnwidth]{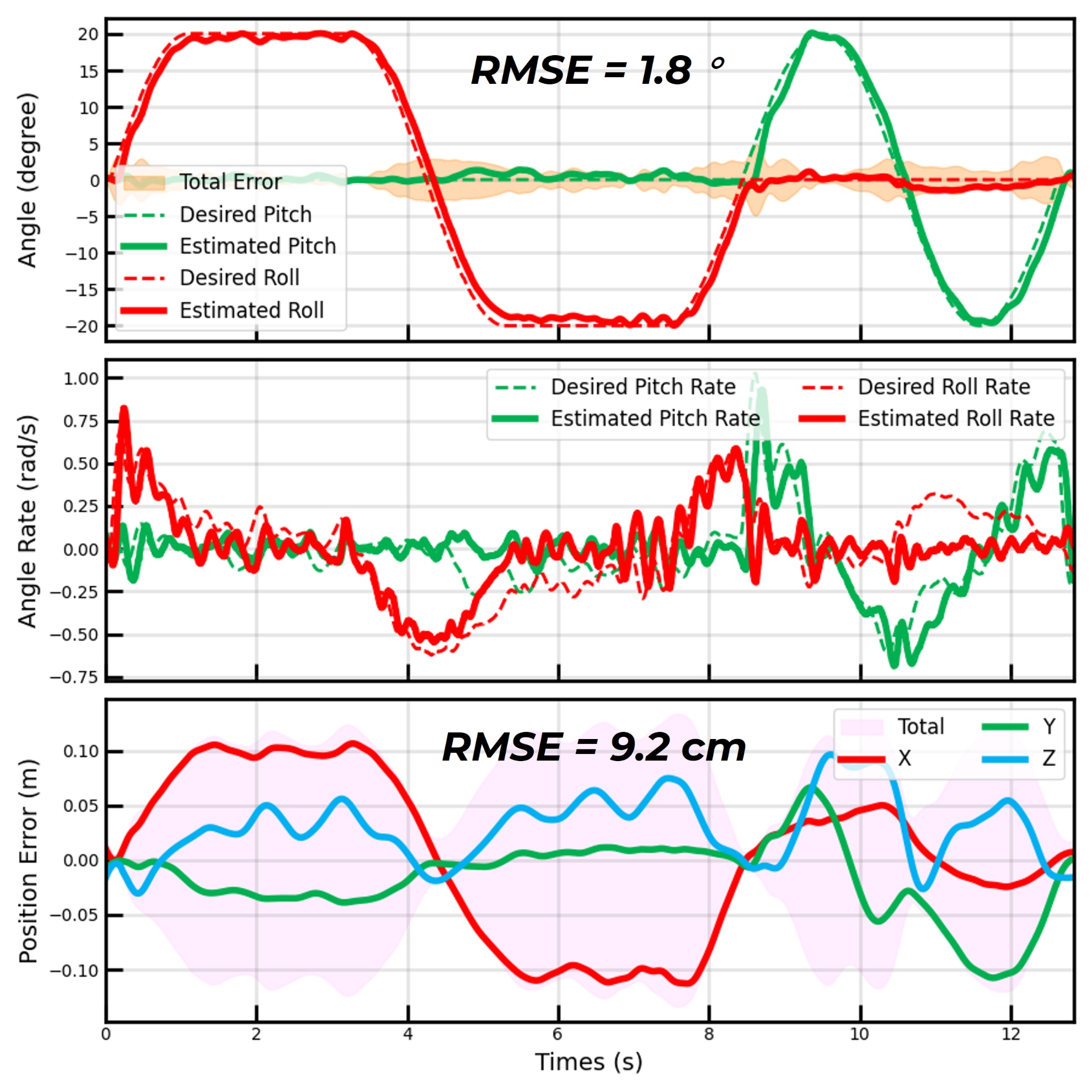}}
	\captionsetup{font={small}}
	\caption{ \label{fig:attitude_track}State curves of FLOAT Drone during the continuous hovering attitude tracking control experiment, including attitude tracking, bodyrate tracking and position tracking.}
	\vspace{-0.2cm} 
\end{figure}

To evaluate the dynamic response performance of the proposed UAV under aggressive attitude tracking scenarios, we conduct an experiment in which the drone tracks continuously varying attitude commands while maintaining hover. The experimental result is shown in Fig.\ref{fig:attitude_track}.

The attitude commands followed a sinusoidal pattern with a variation rate of 1.5 rad/s and a maximum attitude angle of 20 degrees. During the first half of the experiment, the UAV is required to hold at its peak attitudes for $\mathrm{\pi / 2}$ seconds. In the latter half, the attitude commands reverted to a continuous sinusoidal variation. This two-phase design allowed comparative analysis of the UAV’s dynamic response characteristics under both step-like attitude transitions and smooth periodic tracking scenarios.

The experimental results show that the drone achieved a RMSE of 1.8 degrees in attitude tracking, validating the drone's capability to precisely track rapidly changing attitude commands.

\subsection{Traversing an Inclined Narrow Gap}

One of the advantages of the FLOAT Drone over underactuated drones is its ability to maintain a fixed tilt posture, enabling it to navigate through challenging terrains such as narrow inclined spaces. Compared to most fully-actuated drones, the FLOAT Drone is more compact, with a circumscribed circle diameter of only 25 cm and a minimum width of just 18 cm, allowing it to pass through smaller spaces.

This capability is demonstrated in the experiment shown in Fig.\ref{fig:top_figure}(b), where the FLOAT Drone can traverse a 25 cm wide gap inclined at 20 degrees. During the experiment, the drone is controlled in position control mode, where the remote controller sends position and attitude commands to the drone. This experiment highlights the drone’s high control precision and adaptability in complex environments.

\subsection{Pulling and Pushing a Deformable Curtain in Confined Spaces}

Another key advantage of the FLOAT Drone is that it generates minimal lateral airflow during close-proximity operations, thereby reducing interference with the manipulated objects. 

To validate this advantage, we conduct an experiment where the drone is used to push and pull objects, as shown in Fig.\ref{fig:top_figure}(c). In this experiment, we select a curtain as the target object, as it is highly deformable and exhibits noticeable changes when exposed to airflow. 3D-printed hooks are attached to the upper and lower sides of the drone, respectively, to enable interaction with curtains. During the experiment, the drone is controlled in position control mode. It approaches the curtain and performs pulling and pushing actions, with its minimum workspace limited to only 30 cm. By showcasing the drone's safety and precision during close-proximity operations, the experiment confirms its ability to perform delicate tasks in confined spaces.

\subsection{Watering the Flowers via Tilted Hover}

To validate the practicality of the FLOAT Drone, we designed an everyday scenario, specifically watering flowers. For humans, using a watering can to water flowers is a simple and ordinary task, but it's not so easy for drones. To accomplish this task, first, the watering can needs to be tilted at a certain angle to release the water. Second, after tilting, the can must remain stable to ensure the water is precisely poured onto the plants. For underactuated drones, such as quadrotors, completing this task solely on their own is impossible, and additional manipulation devices are required. In contrast, our proposed UAV can effortlessly complete this task.

In this experiment, as shown in Fig.\ref{fig:top_figure}(a), a watering can containing water is mounted on top of the drone. Leveraging its fully actuated characteristics, the drone maintained a tilted hovering posture to pour water from the watering can, achieving the function of watering flowers. Through this experiment, the FLOAT Drone demonstrates remarkable flexibility and practicality in executing specific tasks,  offering new possibilities for drone applications in daily life.

\section{Conclusion}
\vspace{-0.1cm}

In this work, we introduce the FLOAT Drone, a compact fully-actuated coaxial UAV designed to address the challenges of close-proximity aerial operations. By integrating a compact coaxial dual-rotor system and control surfaces, the FLOAT Drone minimizes airflow interference while achieving energy-efficient hovering and precise horizontal force control. Its dual-mode controller enables seamless transitions between fully-actuated and underactuated configurations. Experimental validations demonstrate its superior performance in various close-proximity tasks.

This work advances the field of aerial robotics by providing a potential platform for complex aerial interaction tasks. Future work will focus on enhancing the drone's performance through optimized structural design and a deeper understanding of its aerodynamics, while expanding its application scope to diverse aerial manipulation scenarios.

\section{Acknowledgement}
\vspace{-0.1cm}

We would like to thank Neng Pan and Chengao Li for their valuable help and advice on hardware implementation, and we also thank Weijie Kong for the kind help in polishing the figures.

\newlength{\bibitemsep}\setlength{\bibitemsep}{0.00\baselineskip}
\newlength{\bibparskip}\setlength{\bibparskip}{0pt}
\let\oldthebibliography\thebibliography
\renewcommand\thebibliography[1]{
	\oldthebibliography{#1}
	\setlength{\parskip}{\bibitemsep}
	\setlength{\itemsep}{\bibparskip}
}
\bibliography{references}

\end{document}